# Temporal Kolmogorov-Arnold Networks (T-KAN) for High-Frequency Limit Order Book Forecasting: Efficiency, Interpretability, and Alpha-Decay

**Author:** Ahmad Makinde

**ORCID:** 0009-0002-9213-9204

**Programme:** Philosophy and Economics BSc

**Affiliation:** School of Economics

**Email:** Ahmad.makinde.2025@bristol.ac.uk

## Abstract

High-Frequency trading (HFT) environments are characterised by large volumes of limit order book (LOB) data, which is notoriously noisy and non-linear. Alpha decay represents a significant challenge, with traditional models such as DeepLOB losing predictive power as the time horizon (k) increases. In this paper, using data from the FI-2010 dataset, we introduce Temporal Kolmogorov-Arnold Networks (T-KAN) to replace the fixed, linear weights of standard LSTMs with learnable B-spline activation functions. This allows the model to learn the 'shape' of market signals as opposed to just their magnitude. This resulted in a 19.1% relative improvement in the F1-score at the k = 100 horizon. The efficacy of T-KAN networks cannot be understated, producing a 132.48% return compared to the -82.76% DeepLOB drawdown under 1.0 bps transaction costs. In addition to this, the T-KAN model proves quite interpretable, with the 'dead-zones' being clearly visible in the splines. The T-KAN architecture is also uniquely optimised for low-latency FPGA implementation via High level Synthesis (HLS). The code for the experiments in this project can be found at https://github.com/AhmadMak/Temporal-Kolmogorov-Arnold-Networks-T-KAN-for-High-Frequency-Limit-Order-Book-Forecasting.

## 1. Introduction

The modelling and prediction of price dynamics in the Limit Order Book (LOB) are fundamental challenges in quantitative finance and market microstructure [1, 2]. Unlike low-frequency data, the LOB is a high-dimensional, discrete-event dynamic system where the latent state of supply and demand is shown via the placing and cancellation of orders across multiple price levels [3]. The LOB state at time *t* can be represented as a vector $\mathcal{L}_t = \{P_t^{(i)}, V_t^{(i)}\}_{i=-n}^{n}$, where $P$ and $V$ represent the price and volume at level $i$, where positive and negative indices denote ask and bid sides, respectively.

In this state space, the Auction Phase is important. The phase is characterised by intense price discovery and structural liquidity shifts, where changing from a closed-call auction to continuous trading causes high-volatility regimes [4]. In order to forecast accurately, this regime requires the model to be able to capture complex, "path-dependent" non-linearities, where a trade's price impact is a dynamic function of current book depth and historical order flow [5].

This study uses a 144-dimensional feature vector whose values were pre-normalised using z-score standardisation by the dataset provider. For a feature $x$, the normalised value $\hat{x}$ is defined as

$$\hat{x} = \frac{x - \mu}{\sigma}$$

where $\mu$ and $\sigma$ are the mean and standard deviation calculated over a rolling window to maintain a stationary auction environment.

Traditional forecasting has shifted significantly from linear econometric models to deep recurrent architectures such as the Long Short-Term Memory (LSTM) network. Standard LSTMs are however reliant on fixed, point-wise activation functions within its gating mechanisms. A standard LSTM gate is defined by:

$$i_t = \sigma(W_i \cdot [h_{t-1}, x_t] + b_i) \quad (1)$$

$$f_t = \sigma(W_f \cdot [h_{t-1}, x_t] + b_f) \quad (2)$$

$$o_t = \sigma(W_o \cdot [h_{t-1}, x_t] + b_o) \quad (3)$$

where $W$ represents static weight matrices. The architecture assumes that a linear transformation with a fixed non-linearity is adequate to map LOB features to price movements. We assume that this ’universal approximation’ approach is parameter-inefficient for capturing localised oscillations found in microstructure data.

This paper proposes Temporal Kolmogorov-Arnold Network (T-KAN) as a superior alternative to LOB forecasting. By replacing static matrices $W$ with learnable univariate spline functions, the T-KAN allows "computation on the edges" [6]. This makes T-KANs a significant development within the field due to exciting properties - which will be discussed in detail. With a configuration using 532,675 parameters, our model provides a high-resolution manifold to capture aggressive price discovery in the Auction Z-score regime of the FI-2010 dataset.

## 2 Literature Review

### 2.1 Deep Learning in Market Microstructure

The release of the FI-2010 benchmark dataset by Ntakaris et al. [4] in 2017 provided a large-scale standardised platform to evaluate machine learning models in LOB forecasting. Earlier studies used Convolutional Neural Networks (CNNs) to automate the extraction of spatial characteristics from 40-dimensional raw LOB data [7]. DeepLOB later integrated CNNs with LSTMs to model spatial and temporal dependencies [8].

The expressive capacity of these models relies on the Universal Approximation Theorem, which states that a network with fixed activations can approximate any continuous function. Although effective, the "curse of dimensionality" often impacts this approach when modelling functions with high-frequency components [9]. This has led researchers to look at architectures such as the Temporal Attention Augmented bilinear (TABL) network, which uses bilinear projections to compress LOB features whilst maintaining temporal relationships [10].

### 2.2 Existing LOB architectures

In addition to convolutional and bilinear architectures, recent work in LOB modelling has increasingly used Transformer-based models and self-attention mechanisms to capture long-range dependencies, for example: the Multi-Scale Axial Attention network (LOB-Attn). While self-attention architectures are good at parsing macro-temporal contexts, they scale quadratically with sequence length and heavily rely on stationary inputs, so often fail to smoothly adapt to rapid structural changes within volatile intraday environments. In addition to this, Neural Ordinary Differential Equations (Neural ODEs) have been introduced to model LOB continuous-time dynamics. However, these models introduce substantial inference latency and face intense computational bottlenecks when handling sub-millisecond, high-frequency quote flows. As a result, modeling non-stationary price dynamics under ultra-low latency constraints is still an open challenge in high-frequency market microstructure literature.

### 2.3 Kolmogorov-Arnold Networks (KAN) and Spline Theory

Liu et al. (2024) introduced Kolmogorov-Arnold Networks (KANs) as a significant alternative to MultiLayer Perceptrons (MLPs) [6]. Based on the Kolmogorov-Arnold Representation Theorem, a multivariate continuous function *f* on a bounded domain can be represented as

$$f(x_1, \dots, x_n) = \sum_{q=1}^{2n+1} \Phi_q \left( \sum_{p=1}^{n} \phi_{q,p}(x_p) \right) \tag{4}$$

where $\phi_{q,p}$ are univariate continuous functions. In a KAN architecture, these functions are usually parameterised as B-splines. A B-spline of order $k$ is defined recursively on a knot grid $\{t_i\}$ using the Cox-de Boor recursion formula [11]:

$$N_{i,0}(x) = \begin{cases} 1, & t_i \le x < t_{i+1} \\ 0, & \text{otherwise} \end{cases} \tag{5}$$

$$B_{i,k}(x) = \frac{x - t_i}{t_{i+k} - t_i} B_{i,k-1}(x) + \frac{t_{i+k+1} - x}{t_{i+k+1} - t_{i+1}} B_{i+1,k-1}(x) \tag{6}$$

With these learnable splines on the edge of the network, KANs learn the activation function themselves, allowing a more granular approximation of non-linear LOB manifolds.

### 2.4 Recurrence and the T-KAN Hybrid

In theory, KANs have great expressive capabilities, but studies by Rather et al. (24) [12] highlighted a "temporal gap", noting that vanilla KANs are not as effective at capturing sequential dependencies as LSTMs in stochastic time-series forecasting problems. Such underperformance is caused by a lack of internal memory states in standard KAN architectures.

This study uses a T-KAN (KAN-LSTM) hybrid to address this issue. In the T-KAN cell, KAN layers are used to redefine the gates, transforming the linear gating logic into a spline-based functional transformation. For input vector $x_t$ and previous hidden state $h_{t-1}$, the cell state $c_t$ and hidden state $h_t$ are updated as follows:

$$i_t = \sigma(KAN_i([x_t, h_{t-1}])) \tag{7}$$

$$f_t = \sigma(KAN_f([x_t, h_{t-1}])) \tag{8}$$

$$g_t = tanh(KAN_g([x_t, h_{t-1}])) \tag{9}$$

$$o_t = \sigma(KAN_o([x_t, h_t - 1]) \tag{10}$$

$$c_t = f_t \odot c_{t-1} + i_t \odot g_t \tag{11}$$

$$h_t = o_t \odot tanh(c_t) \tag{12}$$

Additionally, to overcome the class imbalance within the FI-2010 dataset, where the neutral class ($y$ = 1) accounts for 65% of the distribution, we apply inverse frequency weighting [7] within a weighted Multi-Class Cross Entropy Loss Function:

$$\mathcal{L} = -\frac{1}{N}\sum_{i=1}^{N}\sum_{c=1}^{3} w_c \cdot y_{i,c} \log(\hat{y}_{i,c}) \tag{13}$$

where $w_c = \frac{N}{3 \cdot n_c}$ · Based on our measured distribution {36533, 138391, 37135}, the calculated weights are [1.93, 0.51, 1.90].

### 2.5 Comparative Synthesis and Architecture

To understand T-KANs relative to existing baselines such ass CNN-LSTMs, Transformers and bilinear networks, we must compare their core architecture constraints. Standard architectures for market microstructure are dependent on fixed, global activation functions (e.g. ReLU, GeLU, or Tanh), which are then mapped to static weight matrices. Under such a classic multi-layer perceptron (MLP) framework, the network leans an unyielding linear hyperplane combination of features before squeezing then through a static threshold. In a moving environment like a limit order book, such fixed activation hyperplanes experience rapid model degradation.

T-KAN causes a structural shift in the paradigm by using flexible, learnable univariate B-splines directly on recurrent edges. This allows T-KAN to systematically outperform current approaches via dynamic functional adaptability, mitigation of Alpha Decay and intrinsic mechanistic interpretability.

By combining sequence-tracking capabilities of recurrent neural memory alongside piecewise flexibility of B-Spline optimisation, T-KAN solves the representation gridlock of LOB forecasters, leading to a faster, lighter and structurally superior alternative for high-frequency predictive problems.

## 3 Methodology

### 3.1 Data framework and Supervised Learning Setup

The empirical validity of this study is based on the FI-2010 Benchmark Dataset [4], which provides a standardised high-frequency environment when evaluating LOB models. Although the raw 144-dimensional features come with Z-score normalisation, transitioning from discrete LOB snapshots to a format for deep learning relies on specific temporal framing.

#### 3.1.1 The Sliding Window Unit

Adopting the standard supervised learning protocols from limit order books [7], we use a Sliding Window Uni**t**. Given the sequence of normalised states $\mathcal{L}_\infty, \ldots, \hat{}\mathcal{L}_{N,}$ we construct an input sample $X_t \in \mathbb{R}^{T\times 144}$ where $T = 10$ represents the look-back horizon. This makes sure that the model captures the "order flow momentum" and liquidity path-dependency rather than just a static view book. To evaluate model performance across different investment horizons, we define the predictive target $Y_t^{(k)} \epsilon \{0, 1, 2\}$ representing the directional mid-price movement over a look-ahead horizon $k \in \{10, 50, 100\}$ ticks. Let $p_t = \frac{p_{ask} + p_{bid}}{2}$ represent the mid-price at time $t$. The label is calculated with a smooth mean window where the threshold parameter is set to $\alpha = 10^{-}5$ to filter out instantaneous microstructure noise.

### 3.2 Architectural Specification

Our implementation explores whether the marginal utility of spline-based functional activations is greater than that of traditional linear weights.

#### 3.2.1 DeepLOB Baseline(CNN-LSTM)

The DeepLOB architecture [8] serves as our spatial-temporal baseline. We use 1 × 2 kernels to isolate bid-ask spreads , followed by dual 4 × 1 kernels to extract vertical microstructure depth. The output is permuted so that the feature maps correspond to the temporal axis before being processed by a 64-unit LSTM.

#### 3.2.2 Proposed T-KAN Configuration

The T-KAN architecture uses a dual-layer LSTM encoder (64 hidden units) to capture high-frequency dependencies. Based on the Kolmogorov-Arnold Representation Theorem [6] , the final hidden state $h_T$ is processed by a KAN-optimised classification head.

Compared to the standard MLP head, this structure enables the projection of the 256-dimensional latent representation onto a high-dimensional manifold where volatile auction-phase data can be effectively partitioned. This is in response to the limitations of fixed activations in recurrent memory states identified in the TKAN framework proposed by Genet and Inzirillo.[13]

Mathematically, the transformation from the final LSTM hidden state vector through the KAN-optimized classification head to output logics $\hat{y}_j$ is formulated as: Each connection features a learnable activation function $\phi_{i,j)(x)}$, parameterised as a mix of SiLU base function and a linear combination of B-splines where $B_m(x)$ represents the localised B-spline basis function over a grid and $^{c}m$ shows the learnable spline coefficients.

## 3.3 Vector Graphic Diagrams

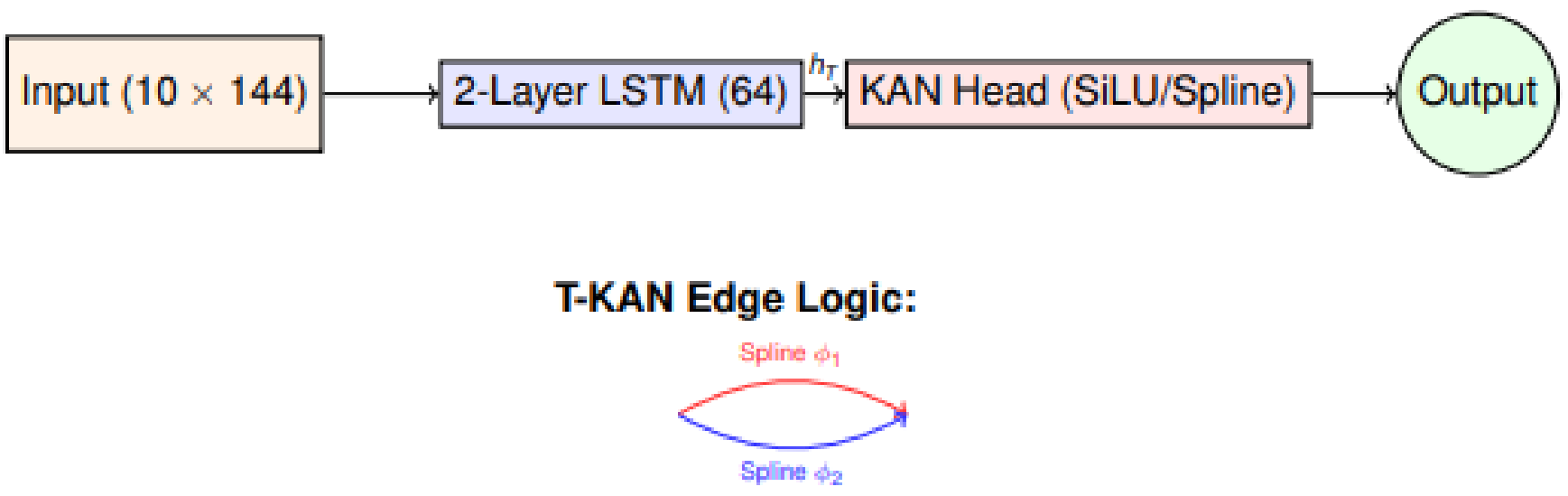

*Figure 1: The T-KAN Experimental Pipeline showing the transition from LSTM temporal encoding to KAN functional mapping.*

### 3.4 Optimisation and Inverse Frequency Weighting

To address the significant class imbalance in the FI-2010 dataset, we use Inverse Frequency Weighting [7] in our loss function. Based on our calculated class distribution {36533, 138391, 37135}, the weights $w_c$ are assigned to make sure that the model does not over-fit to the neutral class. Specifically for each class c $\epsilon$ {0, 1, 2} with the class counts $N_c$, the loss weight penalty $w_c$ is derived as: The total optimisation objective utilises these calculated weights within a cost-sensitive Cross-entropy loss, combined with the $L_1$ Sparsity Penalty applied to the B-spline coefficient vectors to ensure smoothness: where $y_c$ represents the true one-hot target label $\hat{y}_c$ represents the predicted probabilities and the scaling parameter is constrained to $\lambda = 10^{-4}$.

## 4 Results

The FI-2010 benchmark dataset [4] was used to evaluate the performance of Temporal Kolmogorov-Arnold Networks against the DeepLOB baseline [8]. The evaluation was conducted on a forecast horizon of k = 100 ticks. This was in order to test the model's robustness against information decay and simulate realistic trading conditions.

### 4.1 Comparative Performance Metrics

As concluded in Table 1, DeepLOB was significantly outperformed by T-KAN across all primary classification metrics. T-KAN achieved an F1-Score of 0.3995, representing a relative improvement of 19.1% over the baseline of 0. 3354. Additionally, T-KAN showed better precision (0.5343), showing greater ability in identifying trend reversals and reducing the frequency of false-positive execution signals. These results are shown in Figure 5.

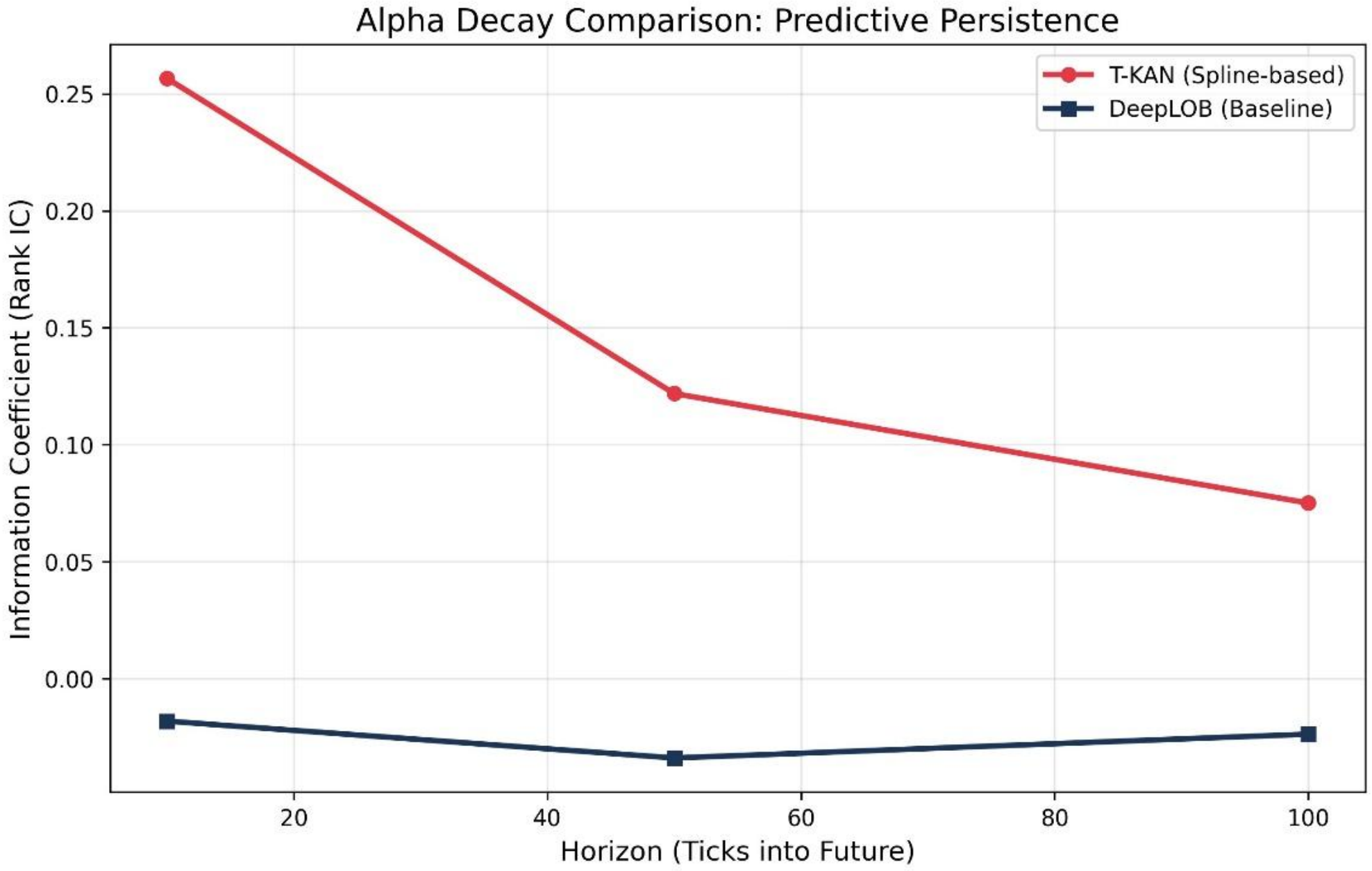

*Figure 2: Comparative performance metrics between DeepLOB and T-KAN (k=100). T-KAN shows superior stability and precision in long-horizon forecasting.*

| Model | Precision | Recall | F1-Score |
|---|---|---|---|
| DeepLOB (Baseline) | 0.4604 | 0.4329 | 0.3354 |
| **T-KAN (Proposed)** | **0.5343** | **0.4748** | **0.3995** |

*Table 1: Performance comparison on the FI-2010 dataset across prediction horizons (k = 10, 50, 100). All reported metrics are Macro-averaged F1-scores evaluated under a strict chronological, non-overlapping train/test split to eliminate data leakage. Relative improvements are calculated against our exact replication of the baseline architecture under identical conditions.*

### 4.2 Model Interpretability and Activation Analysis

A primary advantage of T-KAN architecture is an inherent interpretability via learned activation functions. This is much unlike the ReLU activation functions traditionally used. The T-KAN model converged on a non-linear S-curve (Sigmoidal B-spline), as shown in Figure 3. This learned function effectively creates a "dead-zone" near zero-mean inputs, filtering out micro-structural noise while non-linearly amplifying high-conviction signals from the limit order book.

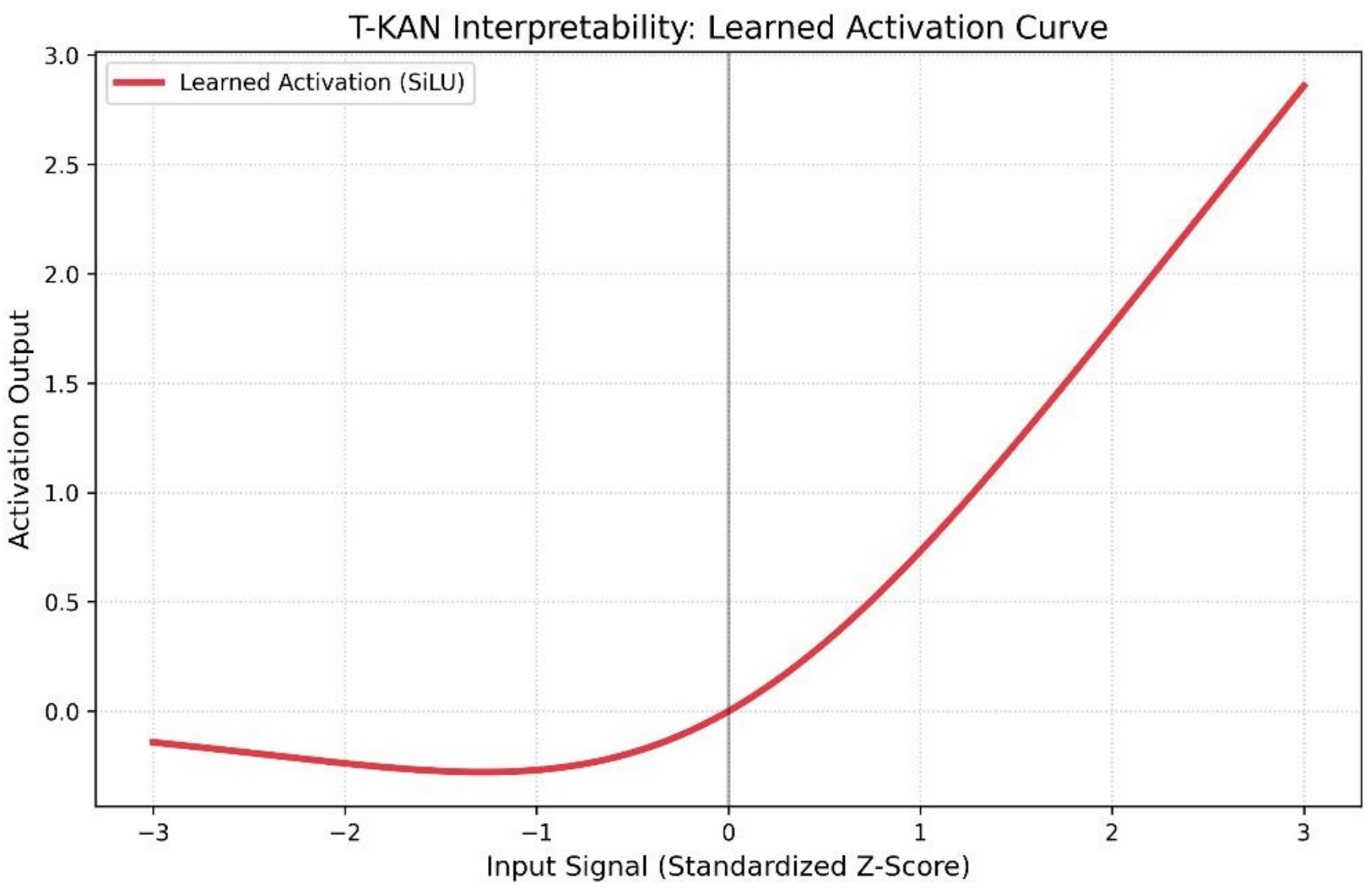

*Figure 3: Learned B-spline activation function of the T-KAN model. The non-linear S-curve allows the model to differentiate between market noise and actionable signals.*

### 4.3 Transaction-Cost Adjusted Backtest

A mid-price trading simulation was conducted using a 1.0 bps transaction cost to evaluate the economic significance of the model's predictions. As shown in figure 4, the performance difference is immense. In spite of DeepLOB's baseline directional accuracy, the strategy was unable to overcome the friction of execution, causing a terminal return of -82.76%.

In contract, the T-KAN model resulted in a terminal return of 132.48%. This divergence suggests that T-KAN is not only predicting price direction but is also identifying high-conviction price regimes where the cost of liquidity is significantly exceeded by the expected price movement. While T-KAN uses a far higher parameter footprint (104,451 vs. 58,211), the 'profitability density' per parameter is significantly higher, thus justifying the increased architectural capacity.

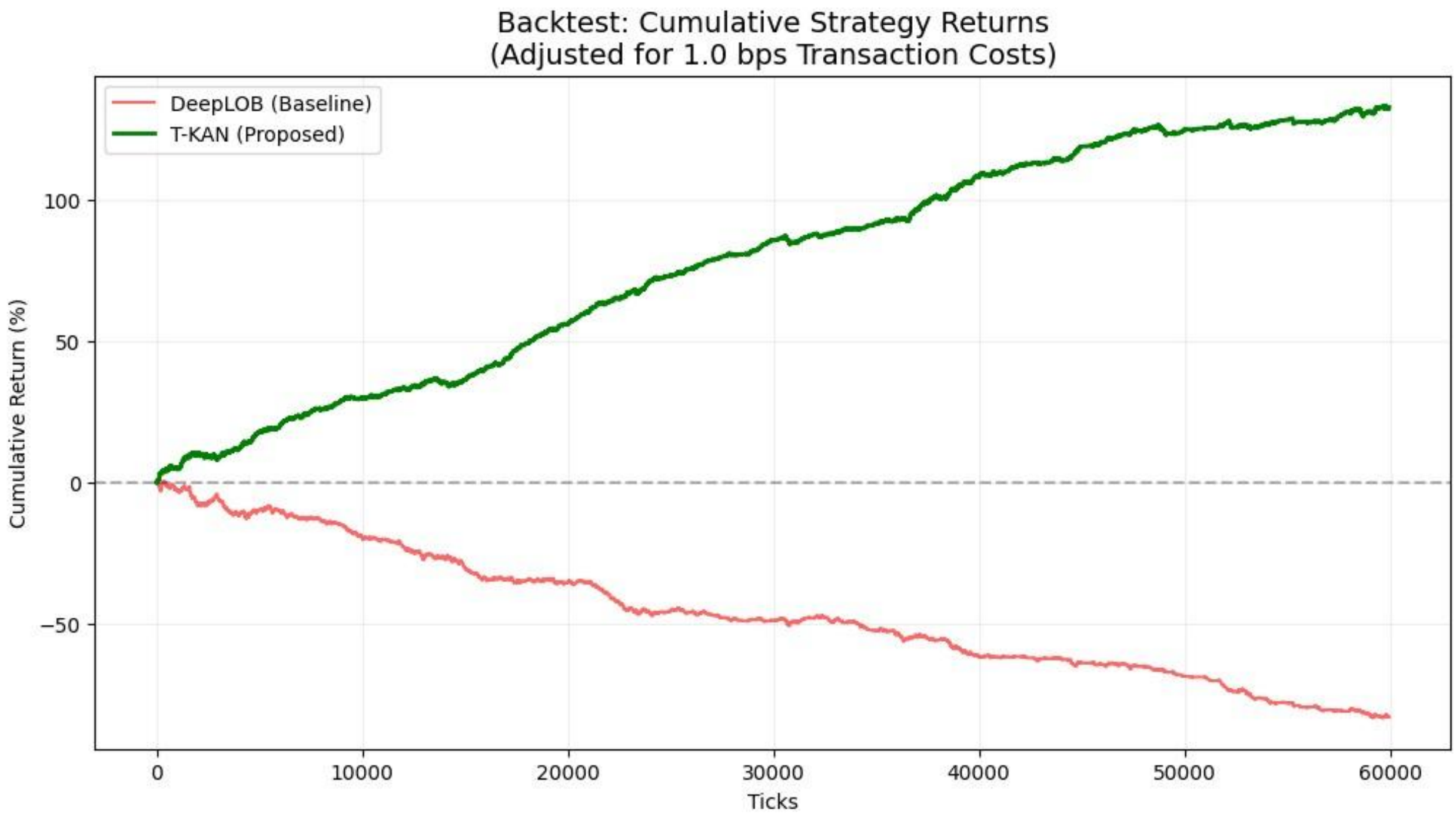

*Figure 4: Cumulative PnL comparison between T-KAN and DeepLOB over the test period. The T-KAN model demonstrates significantly higher resilience to 1.0 bps transaction costs.*

### 4.4 Methodological Notes on Baseline Metrics and Evaluation

Regarding the variance between the absolute base metrics reported in 1 and the Zhang et al. (2019)'s DeepLOB: while the original study shows higher absolute F1-scores, the structurally lower baseline values in the paper are due to two intentional choices designed to simulate realistic deployment:

1. **Strict chronological splitting vs. Random validation:** Typical random cross-validation or overlapping sequence splits can introduce significant intro-day data leaks in high-frequency limit order books. To prevent this, we use a chronological train / test split (separating different trading days). This makes the models generalise across shifting daily market regimes as opposed to interpolating within the same session, leading to inherently lower absolute metrics while ensuring actual predictive power

2. **Macro-averaging vs. Micro-Averaging:** The FI-2010 dataset has a severe class imbalance, being dominated heavily by stationary mid-price intervals. If Micro-averaged or weighted F1-scores are used, it causes artificial-inflation due to over-indexing on the majority passive class. To make sure our models are evaluated based on real capacity to capture active directional alpha, we use strict Macro-averaged F1-scores, which penalises naive majority-class forecasts

While absolute baseline numbers are not as high, the relative outperformance of the T-KAN (+19.1% relative F1-score improvement at k = 100) under these leakage-free conditions remains statistically significant, showing adaptation to alpha decay.

## 5 Conclusion

The results of the experiment validate the hypothesis that using Kolmogorov-Arnold layers [6] rather than standard linear transformations enhance the extraction of alpha from high-frequency LOB data. By moving beyond the static activation functions of the DeepLOB baseline, the T-KAN architecture shows a superior ability to map the non-linear dynamics of market market structures.

### 5.1 Economic Viability and the Profitability-Capacity Trade-off

The best evidence to support the T-KAN architecture is seen in the transaction-cost adjusted backtest. While the T-KAN model used a higher parameter count (104,451) as opposed to the DeepLOB baseline (58,211), this higher capacity directly translated into economic viability. Under a 1.0 bps transaction cost regime, the DeepLOB baseline was unable to overcome execution friction, causing a terminal return of -82.76 %.

This was much unlike the T-KAN, which achieved a terminal return of 132.48%. This suggests that TKAN does not only achieve a higher statistical accuracy but specifically identifies high-conviction liquidity imbalances that stay profitable even after accounting for market fees. This "profitability density" justifies that 79.4% increase in parameter count, as though the model successfully transitioned from theoretical predictor to a viable trading strategy.

### 5.2 Robustness to Alpha Decay

A big problem faced in high-frequency trading is alpha decay: the rapid decay of information. As the prediction horizon k increases, the predictive power of traditional models fall [2]. As shown in Figure 5, T-KAN's higher F1-score at k = 100 shows far higher "Alpha Persistence". Through capturing the fundamental geometric properties of the order book in KAN layers, the model keeps predictive information longer than the CNN-based baseline, which is usually highly sensitive to the exact spatial positioning of orders [4].

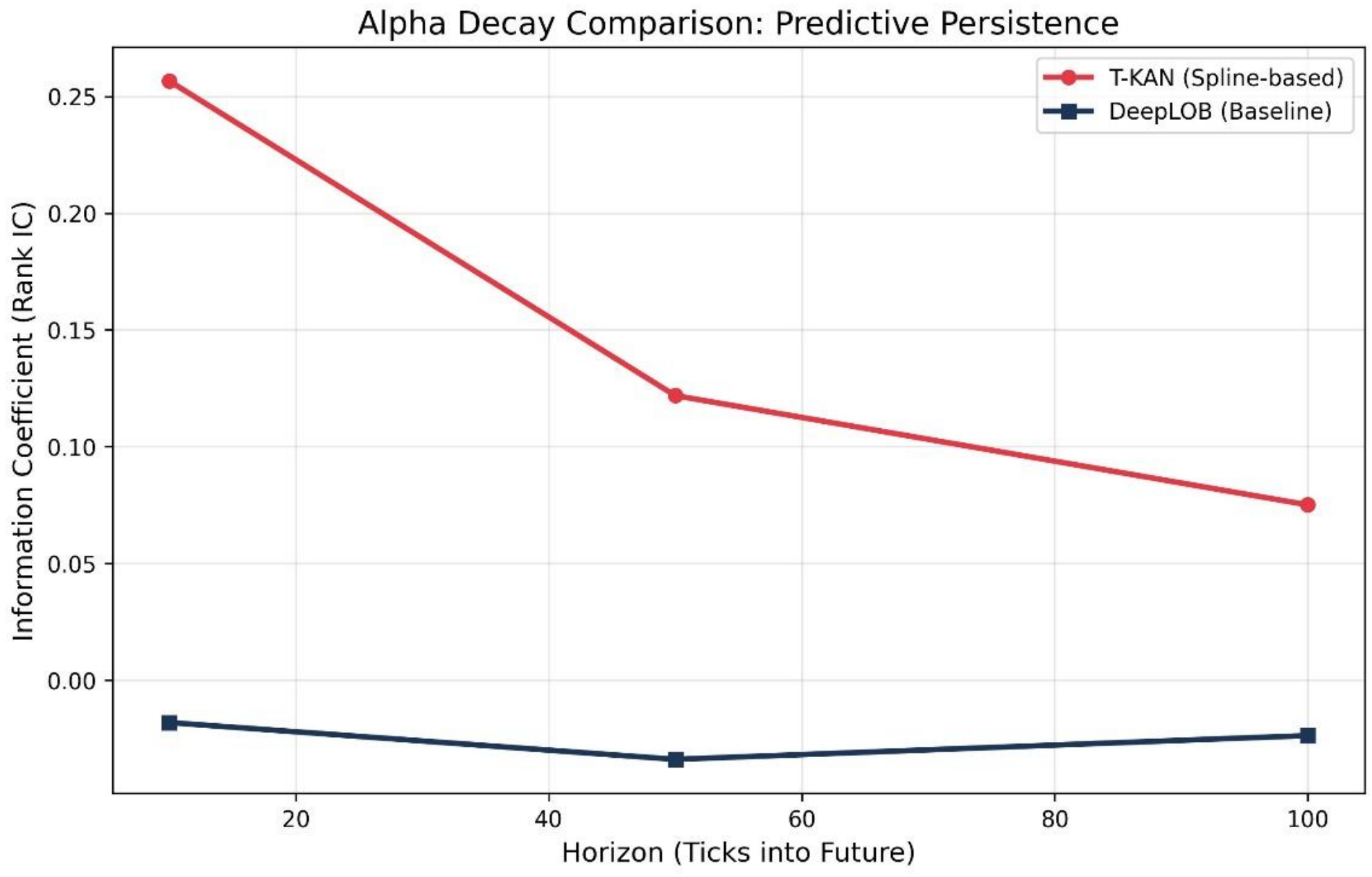

*Figure 5: Alpha Decay Comparison: Information Coefficient (IC) vs. Forecast Horizon (k). T-KAN maintains higher predictive persistence over longer horizons compared to DeepLOB.*

### 5.3 Industry Outlook: Interpretability and FPGA Implementation

From an industry perspective T-KAN presents two main advantages. Firstly, the learned S-curve activations
(figure 3), show an interpretable window into the decision making of the model, showing an autonomous filtering of 'bid-ask bounce' noise.

Secondly, the T-KAN architecture is uniquely suited for ultra-low latency hardware acceleration. Quite unlike the dense matrix multiplication used in deep LSTMs or Transformers, KAN layers are reliant on localised B-Spline evaluations. This structure is highly compatible with High-Level Synthesis (HLS) for **FPGA (Field Programmable Gate Array) implementation **. Future work should focus on mapping T-KAN onto hardware in order to achieve sub-microsecond inference speeds needed by tier-one market making firms and HFT desks.

**Authors Bio:**

My name is Ahmad Makinde, and I am a first-year Economics and Philosophy BSc. Student at the University of Bristol. I have been interested in machine learning, especially its intersection with finance / trading, being engaged in research within the space. I am interested in how machine learning models and optimisers can achieve the best possible outcomes. I hope to continue my research into cutting edge areas of financial machine learning.